\documentclass[letterpaper]{article}
\usepackage[preprint]{aaai2027}
\usepackage[hyphens]{url}
\usepackage{graphicx}
\usepackage{natbib}
\usepackage{caption}
\usepackage{booktabs}
\usepackage{array}
\usepackage{amsmath,amssymb}
\usepackage{multirow}
\newcommand{\method}{NEAR}
\newcommand{\queryarm}{\textsc{Q}}
\newcommand{\dccg}{\textsc{G}}

\newcommand{\best}[1]{\textbf{#1}}
\newcommand{\smallstd}[1]{$\pm$#1}

\title{Beyond Trial Averaging: Anchoring Neural and Visual Representations for Few-Repetition Brain-to-Image Retrieval}
\author{Zhenyao Cui, Siyuan Kan, Dingkun Liu, Dongrui Wu}
\affiliations{School of Artificial Intelligence and Automation, Huazhong University of Science and Technology, Wuhan, China\\
\texttt{zycui@hust.edu.cn}}

\begin{document}
\maketitle

\begin{abstract}
Decoding visual information from brain signals probes neural representations and enables neuro-rehabilitation and dream decoding. Recent brain-to-image retrieval approaches have achieved promising performance, typically by averaging many (up to 80) neural trials per image, requiring repeated stimulus presentation that increases latency, cost, and user burden. When only one or a few repetitions are available, the retrieval accuracy drops sharply. This drop is commonly attributed to query noise because averaging suppresses noise and increases signal stability. However, we find a non-transitive alignment pattern: the low-repetition query signal and the image representation each align with the high-repetition center, but not directly with each other. This pattern shows that query noise is only part of the problem and that gallery placement also affects retrieval. We therefore propose a neural-anchor-based retrieval (NEAR) framework that treats the high-repetition center as an anchor and approaches it from both sides: a denoiser pulls the noisy query toward the true anchor, and a small network predicts each candidate's pseudo anchor from its image and pulls the image toward it. Across four datasets spanning EEG, MEG and fMRI, NEAR consistently improved retrieval in the few-repetition regime. On THINGS-EEG2, it improved 200-way Top-1 accuracy by 5.7 and 9.3 percentage points respectively, when averaging one and four repetitions. By anchoring neural and visual representations, NEAR reduces reliance on repeated acquisition and brings neural retrieval closer to real-world deployment.
\end{abstract}

\section{Introduction}
Brain-to-image retrieval identifies the image a person is viewing from the corresponding neural response. It can probe visual representations and support neural rehabilitation and dream decoding \citep{kay2008,chaudhary2016,horikawa2013}. Modern systems map EEG, MEG, or fMRI signals and candidate images to a shared space, then retrieve the image most similar to the neural query. Advances in neural encoders, training objectives, and visual targets have produced strong benchmark results \citep{nice,atms,muse,neuroclip,hcf}.

These systems, however, often average repeated neural measurements of the same image to suppress trial-specific noise and stabilize the query across EEG, MEG, and fMRI \citep{giesbrecht2024,thingsmeg,nsd}. THINGS-EEG2, for example, presents each test image 80 times before evaluation \citep{thingseeg2,nice}. This improves reliability but also increases latency, acquisition cost, and participant burden \citep{scalinglaws}.

\begin{figure}[t]
\centering
\includegraphics[width=\linewidth]{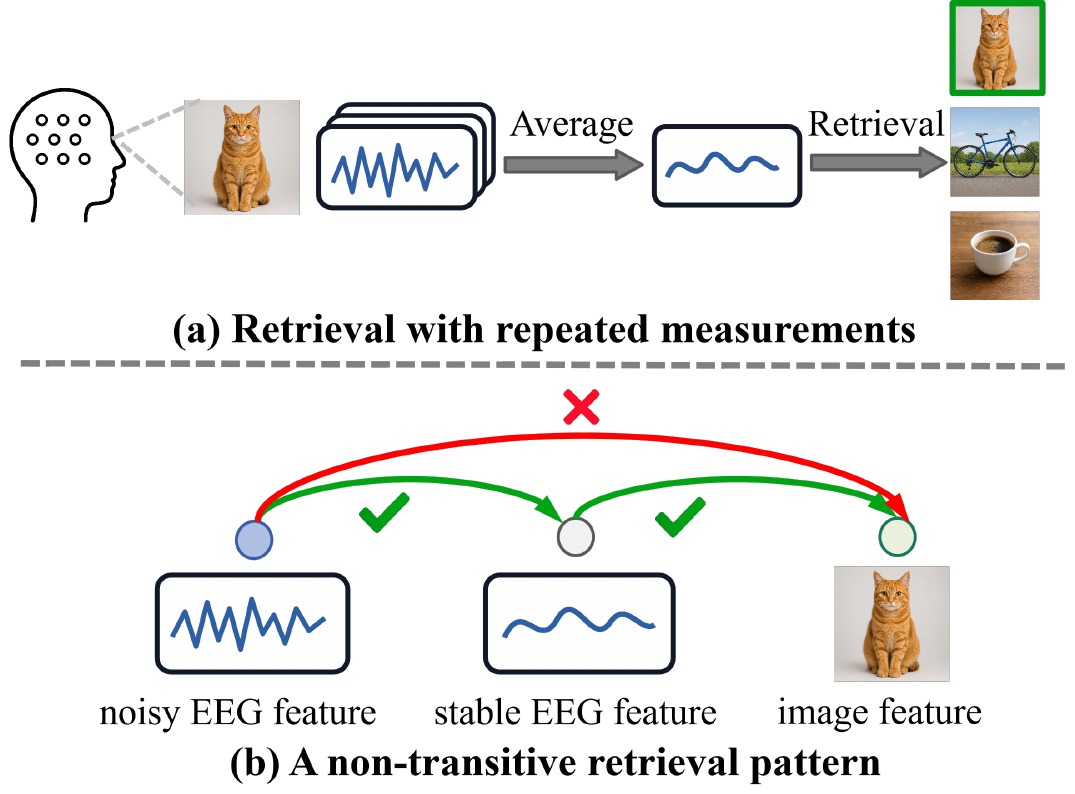}
\caption{Non-transitive few-repetition retrieval.
(a) Averaging repeated measurements produces a stable neural representation that retrieves the viewed image.
(b) The few-repetition query and image each align with this representation, but not reliably with each other.}
\label{fig:intro-scene}
\end{figure}

This dependence becomes problematic when only a few repetitions are available \citep{scalinglaws}. On THINGS-EEG2, the same retriever's $200$-way Top-1 accuracy falls from $83.4\%$ with $80$ repetitions to $14.0\%$ with one. The standard explanation is that a query from one or a few repetitions contains substantial trial-specific noise, whereas averaging reveals the stable image-evoked response \citep{single_trial_bci,xdawn,rerp,eeg2erp}.

However, we find that the query is only part of this problem. We design a controlled target-swap experiment and compare three paths for the same concepts: from a few-repetition query to the stable representation estimated from many repetitions of the same image, from this stable representation to the image, and directly from the query to the image. The experiment reveals the non-transitive pattern in Figure~\ref{fig:intro-scene}: the few-repetition query and image representation each align with the stable representation, but align less well with each other. This pattern suggests improving retrieval across repetition counts by using the stable neural representation as an anchor and moving both the query and image toward it.

Therefore, we propose neural-anchor-based retrieval (\method{}), which uses the stable neural representation as a common anchor through two branches. Query Anchoring denoises the few-repetition query toward its neural anchor. Gallery Anchoring predicts a pseudo anchor for each candidate image and combines it with the original visual representation during scoring.

Across four datasets spanning EEG, MEG, and fMRI, \method{} improves each base retriever despite differences in signal type, candidate-set size, and neural encoder. On THINGS-EEG2, it raises $200$-way Top-1 accuracy by $5.7$ and $9.3$ percentage points with one and four repetitions, respectively, and improves all ten participants at both budgets. These results show that neural anchoring transfers across modalities and improves few-repetition retrieval.

\noindent\textbf{Contributions.}
\begin{itemize}
    \item We reveal a non-transitive pattern that identifies the stable neural representation as an intermediate anchor between a few-repetition query and its image.
    \item We propose \method{}, which uses this anchor on both sides of retrieval through Query Anchoring and Gallery Anchoring. Query Anchoring denoises the few-repetition query, while Gallery Anchoring predicts a pseudo anchor for each candidate image.
    \item Across four datasets spanning EEG, MEG, and fMRI, \method{} improves few-repetition retrieval. Further analyses show that Gallery Anchoring improves gallery placement and complements different query estimators.
\end{itemize}

\section{Related Work}
\paragraph{Brain-to-image retrieval with repeated measurements.}
Most brain-to-image retrieval methods align image representations with neural features averaged over repeated measurements. EEG systems encode the resulting stimulus-locked response in a pretrained visual space and rank held-out images by cross-modal similarity. THINGS-EEG2 averages up to $80$ test responses per image, while Alljoined evaluates the same setting under lower-SNR EEG \citep{thingseeg2,alljoined16m}. NICE, ATM-S, MUSE, NeuroCLIP, HCF, and UBP improve the neural encoder, training objective, or visual target within this setting \citep{nice,atms,muse,neuroclip,hcf,ubp}. Related fMRI systems also use repeated presentations or averaged response estimates to stabilize the query \citep{mindeye,mindeye2,efird,bok}. Banville~et~al.\ systematically vary the number of averaged responses across EEG, MEG, and fMRI datasets and show consistent but diminishing gains \citep{scalinglaws}. The retrieval studies above demonstrate the gains from repeated averaging, while Banville~et~al.\ quantify their diminishing returns. Our work likewise studies few-repetition retrieval, but further investigates why its accuracy degrades and proposes a method to improve it.

\paragraph{Few-repetition neural-signal estimation and denoising.}
Across EEG, MEG, and fMRI, repeated averaging aims to recover a stable stimulus-linked response from noisy measurements. EEG and MEG average time-locked trials to suppress ongoing activity and sensor noise, while fMRI combines responses from repeated presentations to stabilize voxel-response estimates \citep{single_trial_bci,thingsmeg,nsd}. When only a few repetitions are available, denoising methods pursue the same goal without extensive averaging. Spatial filters reweight channels to emphasize event-related components. Regression-based estimators and robust averaging recover the shared response while limiting nuisance variation. Learned denoisers predict the common signal from independently noisy recordings \citep{makeigica,xdawn,rerp,eeg2erp,noise2noise}. These methods motivate \method{}'s query branch. Unlike these denoising methods, we identify a broader pattern in brain-to-image retrieval: few-repetition accuracy depends jointly on the query and gallery. Gallery-side correction is therefore needed alongside query denoising.

\paragraph{Stimulus-to-brain encoding models.}
Stimulus-to-brain encoding models provide the closest precedent for predicting neural references from stimuli. Kay~et~al.\ predict candidate voxel patterns for fMRI identification, Gifford~et~al.\ map image features to channel-by-time EEG responses, and Concept2Brain synthesizes multi-channel EEG through an electrophysiological latent space \citep{kay2008,thingseeg2,concept2brain}. These studies show that moving stimuli toward neural space can improve decoding, although they primarily predict responses in the original measurement space. \method{} focuses on retrieval with fewer repetitions. Guided by the non-transitive analysis above, it moves both the query and gallery toward the same stable neural anchor in the learned retrieval space.

\section{Neural-Anchor-Based Retrieval}
\label{sec:trial-axis}
\label{sec:method}

\begin{figure}[t]
\centering
\includegraphics[width=\linewidth]{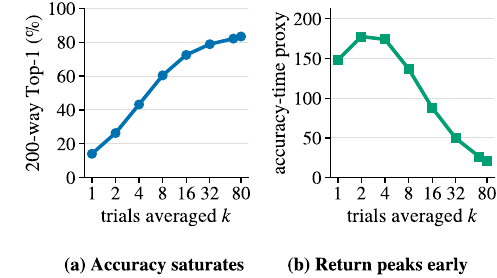}
\caption{Accuracy and acquisition return across repetitions.
(a) Accuracy rises quickly and then approaches a plateau.
(b) Stimulus-time-normalized ITR peaks at $k=2$ because later gains no longer offset added stimulus time.}
\label{fig:problem}
\end{figure}

\paragraph{Problem setup.}
For concept $c$, let $x_{c,i}$ denote the neural measurement at repetition $i$, let $n_c$ denote the number of repetitions of that concept available for training, and let $v_c$ denote its frozen visual feature. We define the measurement averaged over exactly $k$ repetitions and the resulting noisy query as
\begin{equation}
x^{(k)}_c
=
\frac{1}{k}\sum_{i=1}^{k}x_{c,i},
\qquad
q_c^{(k)}
=
f_\theta\!\left(x^{(k)}_c\right).
\label{eq:query}
\end{equation}
We reserve $\bar q_c$ for a stable neural anchor estimated from the larger reference set available in the current split. The target-swap analysis uses repetitions disjoint from the $k$ query repetitions, while Gallery Anchoring uses all available training repetitions. All neural, visual, and pseudo-anchor embeddings are $\ell_2$-normalized before scoring. The standard retriever ranks candidates by $\left(q_c^{(k)}\right)^\top v_j$ and returns the highest-scoring image. Throughout, $j$ indexes candidates in the gallery; when a competitor is meant we write $j\ne c$. Existing systems stabilize this query through extensive averaging. We study the same task with few repetitions.

\paragraph{Few-repetition accuracy and acquisition return.}
We first fix the retriever, preprocessing, and gallery and vary only $k$ to observe how Top-1 accuracy changes. Figure~\ref{fig:problem}a shows that accuracy rises quickly and then approaches a plateau. Each repetition, however, requires another stimulus presentation, so accuracy alone cannot capture the practical trade-off between retrieval performance and acquisition cost. We also use the standard Wolpaw information-transfer rate (ITR), which combines Top-1 accuracy among $N$ candidates with total stimulus time to report information transferred per minute \citep{wolpaw2002,mcfarland2003}. Figure~\ref{fig:problem}b shows that acquisition efficiency peaks at $k=2$ and then declines because later gains no longer offset the added stimulus time. Thus, although accuracy remains low with few repetitions, this regime offers the highest return per unit acquisition time and is the most relevant range for practical improvement.

\subsection{Analyzing the Retrieval Gap}
\label{sec:retrieval-analysis}

\paragraph{Controlled target swap.}
Test-time averaging improves signal-to-noise ratio and decoding performance, with diminishing returns as more repetitions are added \citep{scalinglaws}. This naturally suggests an unstable query as the main source of the low-repetition decline. However, the repetition curve alone does not show how much stable stimulus information a few-repetition query already contains. We therefore construct a few-repetition query $q_c^{(k)}$ and a stable neural representation $\bar q_c$ from disjoint repetitions of the same image. Most prior work effectively evaluates $\bar q_c{\to}v_c$: it averages many repetitions to obtain a stable query and retrieves the corresponding image from the visual gallery. We retain this conventional path, introduce $q_c^{(k)}{\to}\bar q_c$, and evaluate $q_c^{(k)}{\to}v_c$ as the direct few-repetition path. All three comparisons use the same candidate concepts and gallery size. Only the query or target representation changes.

\begin{figure}[t]
\centering
\includegraphics[width=\linewidth]{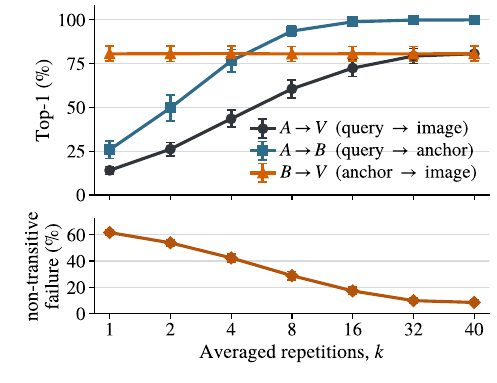}
\caption{Target swap exposes non-transitive retrieval.
With $q_c^{(k)}$ and $\bar q_c$ estimated from disjoint repetitions, $q_c^{(k)}$ retrieves $\bar q_c$ more accurately than its image $v_c$ (top). Direct retrieval can still fail when $q^{(k)}{\to}\bar q$ and $\bar q{\to}v$ both succeed (bottom).}
\label{fig:target-swap}
\end{figure}

\paragraph{Stable neural information emerges early.}
Figure~\ref{fig:target-swap} shows how the accuracies of the three target-swap paths change with $k$. The $q^{(k)}{\to}\bar q$ path rises from $25.8\%$ at $k=1$ to $76.5\%$ at $k=4$ and $93.6\%$ at $k=8$, showing that a few-repetition query already contains stable stimulus information. At $k=4$, $\bar q{\to}v$ and $q^{(k)}{\to}\bar q$ reach $80.7\%$ and $76.5\%$, while direct $q^{(k)}{\to}v$ reaches only $43.5\%$. This gap also appears within the same cases: when both paths through $\bar q$ succeed, direct retrieval still fails in $42.4\%$ of them. We call this phenomenon a non-transitive pattern: a few-repetition query can retrieve its stable neural representation, and that representation can retrieve the image, while direct query-to-image retrieval remains unreliable. To exclude an advantage from comparing two neural representations, we keep the anchor gallery fixed and permute the concept--anchor correspondence. Accuracy drops to $0.1\%$, confirming that this path relies on the correct concept-specific anchor. This intermediate connection suggests using $\bar q_c$ as a common anchor for the few-repetition query and candidate image.

\subsection{A Two-Term Retrieval Condition}
\label{sec:error-analysis}

Let $c$ be the correct concept and $j\ne c$ a competitor, with normalized visual features $v_c$ and $v_j$. Because cosine similarity is their dot product, $c$ ranks above $j$ when $\left(q_c^{(k)}\right)^\top(v_c-v_j)>0$. Let $\epsilon_c^{(k)}=q_c^{(k)}-\bar q_c$ denote the query error after averaging $k$ repetitions. Substitution separates the score advantage into two terms:
\begin{equation}
\left(q_c^{(k)}\right)^\top(v_c-v_j)
=\bar q_c^\top(v_c-v_j)
+\left(\epsilon_c^{(k)}\right)^\top(v_c-v_j).
\label{eq:margin-decomposition}
\end{equation}
We call the first term $\gamma_{cj}=\bar q_c^\top(v_c-v_j)$ the stable gallery margin: the correct image's advantage over $j$ when the query is exactly at its stable anchor. The second term is the amount by which the remaining query error raises or lowers this advantage. Thus, retrieval depends on both the gallery margin and the query error.

\begin{figure*}[t]
\centering
\includegraphics[width=\textwidth]{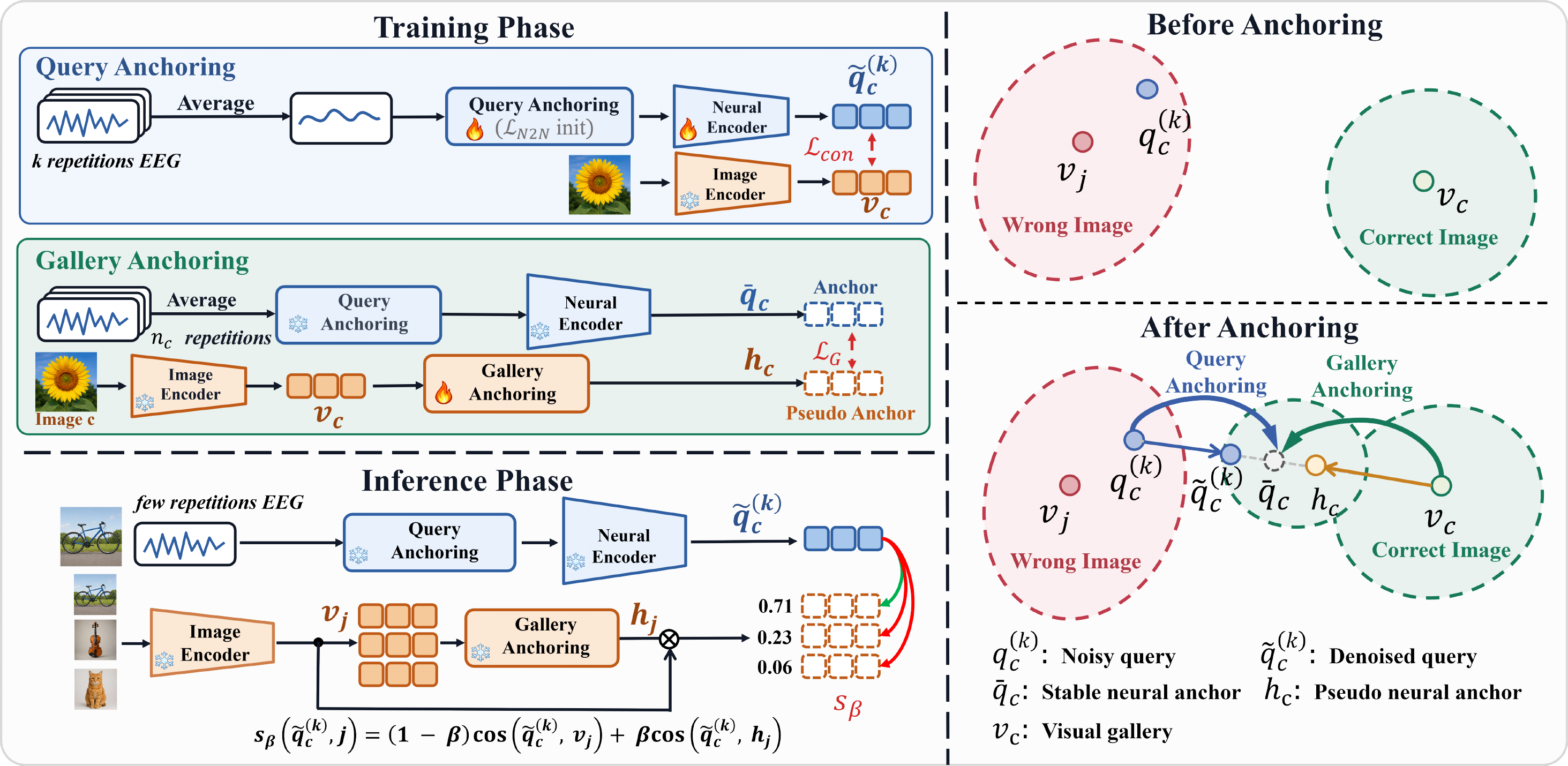}
\caption{Training and inference in \method{}. Query Anchoring is learned from $k$-repetition averages and then frozen; Gallery Anchoring is learned from image--anchor pairs, using all $n_c$ training repetitions of concept $c$ to form the anchor $\bar q_c$. At inference the two branches give a denoised query and a pseudo anchor per candidate, and $\otimes$ combines the two cosine similarities, not the two representations, as in \eqref{eq:anchored-score}. The right panels illustrate both effects.}
\label{fig:overview}
\end{figure*}

To determine how much query error the gallery can tolerate, we consider the most harmful error direction. The Cauchy--Schwarz inequality \citep{boyd2004convex} bounds the magnitude of the second term:
\[
\begin{aligned}
\left|
\left(\epsilon_c^{(k)}\right)^\top(v_c-v_j)
\right|
\;&\le
\left\|\epsilon_c^{(k)}\right\|_2
\left\|v_c-v_j\right\|_2,\\
\left(q_c^{(k)}\right)^\top(v_c-v_j)
\;&\ge
\gamma_{cj}
-
\left\|\epsilon_c^{(k)}\right\|_2
\left\|v_c-v_j\right\|_2.
\end{aligned}
\]
Therefore, a sufficient condition for the correct image to remain above every competitor is
\begin{equation}
\left\|\epsilon_c^{(k)}\right\|_2 < \rho_c,
\qquad
\rho_c =
\min_{j\ne c}
\frac{\gamma_{cj}}{\left\|v_c-v_j\right\|_2}.
\label{eq:basinradius}
\end{equation}
For each competitor, this ratio is the distance from $\bar q_c$ to the decision boundary where $j$ overtakes $c$. Their minimum, $\rho_c$, is the gallery tolerance. Any query error smaller than a positive $\rho_c$ is guaranteed to preserve every ranking, although larger errors may still succeed because the bound uses the worst direction. A non-positive $\rho_c$ means that moving the query to its stable anchor alone cannot guarantee correction.

This condition also explains how the two terms interact across repetition counts. At very low $k$, large query error dominates the ranking. As averaging stabilizes the query, more cases approach the decision boundary, where a larger gallery tolerance can move them to the correct side. At high $k$, the query is already precise, but Gallery Anchoring can still recover concepts with small or non-positive stable margins.

The two terms directly motivate \method{}. Query Anchoring replaces $q_c^{(k)}$ with the denoised query $\tilde q_c^{(k)}$ to reduce its error from $\bar q_c$. Gallery Anchoring predicts a pseudo anchor $h_j$ for each candidate and uses it to increase the stable gallery margin. The two branches therefore target the query error and gallery tolerance $\rho_c$. Figure~\ref{fig:overview} summarizes their training and inference.

\subsection{Query Anchoring}
\label{sec:querylever}
Query Anchoring denoises the measured signals before neural encoding. We use Noise2Noise as the denoising module \citep{noise2noise}. Consider two repetitions of the same training image, $x_{c,a}=u_c+n_a$ and $x_{c,b}=u_c+n_b$, where $u_c$ is their shared stimulus-linked response and the two zero-mean noise terms are independent. We train the denoiser $a_\phi$ to predict one repetition from the other:
\begin{equation}
\mathcal{L}_{\mathrm{N2N}}
=
\left\|a_\phi(x_{c,a})-x_{c,b}\right\|_2^2.
\label{eq:n2n-loss}
\end{equation}
Because the target noise $n_b$ is independent of the input, it cannot be predicted consistently and averages out across training pairs. The expected squared loss therefore retains the shared response $u_c$ without requiring a clean target \citep{noise2noise}.

After this initialization, we train $a_\phi$ and the neural encoder $f_\theta$ on averages with different repetition counts using a standard symmetric contrastive retrieval loss $\mathcal{L}_{\mathrm{con}}$ \citep{clip}. Its two directions bring matching neural and image embeddings closer while separating mismatched pairs. At inference, we average the $k$ repetitions, apply $a_\phi$, and encode the denoised signal with $f_\theta$ to obtain $\tilde q_c^{(k)}$.

For datasets whose training SNR is too low to train Noise2Noise reliably, we instead use fixed wavelet thresholding. We apply the fixed operator $w$ to each repetition, average the filtered signals, and encode them with $f_\theta$. This pathway uses the same contrastive loss while keeping $w$ fixed. Both pathways accept any repetition count because averaging preserves the signal shape. Without Query Anchoring, $\tilde q_c^{(k)}=q_c^{(k)}$. After training either query pathway, we freeze it before training Gallery Anchoring.

\subsection{Gallery Anchoring and Retrieval}
\label{sec:visualrecenter}
We encode the average of the $n_c$ training repetitions as $\bar q_c=\tilde q_c^{(n_c)}$, the most stable neural target available for concept $c$. When Query Anchoring is absent, $\tilde q_c^{(k)}=q_c^{(k)}$.

Then, we freeze the neural and visual encoders and train an image-to-neural predictor $g_\psi$ on these targets:
\begin{equation}
h_c
=
g_\psi(v_c),
\qquad
\mathcal{L}_{\mathrm{G}}
=
1-\cos\!\left(h_c,\bar q_c\right).
\label{eq:gallery-loss}
\end{equation}
Cosine regression directly fits each candidate reference to the stable neural target identified by the analysis.
For an unseen candidate $j$, $h_j=g_\psi(v_j)$ predicts its unavailable $\bar q_j$ from the image alone. We call $h_j$ a pseudo anchor.

However, a pseudo anchor can contain prediction error, while the original visual feature retains useful semantic information. We therefore score each candidate with both references:
\begin{equation}
s_\beta\!\left(\tilde q_c^{(k)},j\right)
=
(1-\beta)\cos\!\left(\tilde q_c^{(k)},v_j\right)
+
\beta\cos\!\left(\tilde q_c^{(k)},h_j\right).
\label{eq:anchored-score}
\end{equation}
Here $\beta\in[0,1]$ balances the two similarities. Setting $\beta=0$ gives standard visual-gallery scoring, while $\beta=1$ uses only pseudo anchors. At the stable anchor, \eqref{eq:anchored-score} changes the gallery term in \eqref{eq:margin-decomposition} from $\gamma_{cj}$ to $(1-\beta)\gamma_{cj}+\beta\bar q_c^\top(h_c-h_j)$. Training $h_c$ toward $\bar q_c$ makes the new term favor the correct candidate, increasing the gallery tolerance $\rho_c$. Query Anchoring instead reduces the query error. The signed-margin analysis in Section~\ref{sec:fmri} tests this effect directly.

\section{Experiments}
\label{sec:experiments}
\subsection{Experimental Setup}
Our primary evaluation uses two 200-way within-subject EEG benchmarks. THINGS-EEG2 has 10 participants and multiple training and test repetitions, while Alljoined-1.6M has 20 participants and lower-SNR EEG \citep{thingseeg2,alljoined16m}. Because both provide multiple training repetitions, we evaluate Query and Gallery Anchoring together across repetition counts. For these EEG datasets, repetitions of the training images provide stable targets for the gallery predictor, which receives only candidate image representations at inference.

Besides these EEG benchmarks, we use THINGS-MEG and NSD. THINGS-MEG has multiple test repetitions but one training repetition per stimulus, so we keep its query pathway fixed and evaluate Gallery Anchoring alone \citep{thingsmeg}. The published NSD decoder already provides a usable query from individual 7T fMRI repetitions, so we also keep this pathway fixed and use NSD to isolate Gallery Anchoring. For each participant, we use 8,018 non-test images to fit the query decoder and gallery predictor, with 1,000 additional non-test images used to select the query decoder. The 982 images shared across participants and all their fMRI repetitions are reserved for retrieval evaluation \citep{nsd,efird,bok}.

For both EEG benchmarks, we follow HCF's preprocessing, neural encoder, visual head, and within-subject protocol \citep{hcf}. THINGS-EEG2 uses the Noise2Noise-initialized, repetition-count-aware query pathway. Alljoined has lower training SNR, and matched comparisons favor fixed wavelet thresholding in this regime. We apply it to each repetition before averaging.

THINGS-MEG follows its published preprocessing and within-subject retriever \citep{thingsmeg}. For NSD, we follow the global-vector setting of Efird et al.\ and evaluate the CLIP and robust FARE targets used by Bok et al.\ \citep{efird,bok}. By default, $g_\psi$ is an MLP with 2,048 units in each hidden layer. Each comparison changes only the tested component, and we fix $\beta=0.75$ across all datasets and repetition counts. Tables and figures use Base to denote the corresponding unanchored retriever. Unless stated otherwise, results average three fixed seeds, and standard deviations are reported where available. We report Top-1 accuracy throughout.

\subsection{Main Results on EEG}

\begin{table*}[!t]
\centering\rmfamily\small\setlength{\tabcolsep}{3.0pt}
\begin{tabular*}{\textwidth}{@{\extracolsep{\fill}}ccccccccc@{}}
\toprule
Dataset & Method & $k{=}1$ & $4$ & $8$ & $16$ & $32$ & $64$ & $80$ \\
\midrule
\multirow{4}{*}{THINGS-EEG2}
 & Base             & 14.0\smallstd{0.00} & 43.2\smallstd{0.10} & 60.4\smallstd{0.30} & 72.5\smallstd{0.30} & 78.8\smallstd{0.30} & 82.1\smallstd{0.10} & 83.4\smallstd{0.30} \\
 & $+$\queryarm{}   & 18.4\smallstd{0.10} & 47.5\smallstd{0.40} & 62.1\smallstd{0.40} & 71.7\smallstd{0.30} & 77.2\smallstd{0.40} & 79.9\smallstd{0.50} & 80.5\smallstd{0.30} \\
 & $+$\dccg{}       & 14.9\smallstd{0.10} & 47.0\smallstd{0.10} & 65.6\smallstd{0.20} & 77.9\smallstd{0.30} & 84.1\smallstd{0.20} & \best{87.5}\smallstd{0.30} & \best{88.0}\smallstd{0.10} \\
 & \method{}        & \best{19.8}\smallstd{0.10} & \best{52.5}\smallstd{0.20} & \best{68.5}\smallstd{0.20} & \best{79.0}\smallstd{0.20} & \best{84.4}\smallstd{0.20} & 87.1\smallstd{0.50} & 87.6\smallstd{0.40} \\
\midrule
\multirow{4}{*}{Alljoined}
 & Base             & 1.41\smallstd{0.01} & 2.85\smallstd{0.07} & 4.41\smallstd{0.10} & 6.81\smallstd{0.10} & 9.57\smallstd{0.30} & 12.93\smallstd{0.47} & 14.30\smallstd{0.57} \\
 & $+$\queryarm{}   & 1.79\smallstd{0.00} & 3.73\smallstd{0.04} & 5.65\smallstd{0.05} & 8.42\smallstd{0.04} & 11.68\smallstd{0.15} & 15.56\smallstd{0.21} & 16.72\smallstd{0.06} \\
 & $+$\dccg{}       & 1.51\smallstd{0.01} & 3.17\smallstd{0.08} & 4.92\smallstd{0.07} & 7.52\smallstd{0.14} & 10.60\smallstd{0.29} & 14.18\smallstd{0.16} & 15.69\smallstd{0.52} \\
 & \method{}        & \best{1.98}\smallstd{0.01} & \best{4.29}\smallstd{0.05} & \best{6.64}\smallstd{0.04} & \best{9.79}\smallstd{0.10} & \best{13.60}\smallstd{0.23} & \best{17.72}\smallstd{0.29} & \best{18.84}\smallstd{0.25} \\
\bottomrule
\end{tabular*}
\normalsize
\caption{Top-1 accuracy (\%) on THINGS-EEG2 (10 participants) and Alljoined (20 participants). Bold marks the best result at each repetition count.}
\label{tab:main_eeg}
\includegraphics[width=\textwidth]{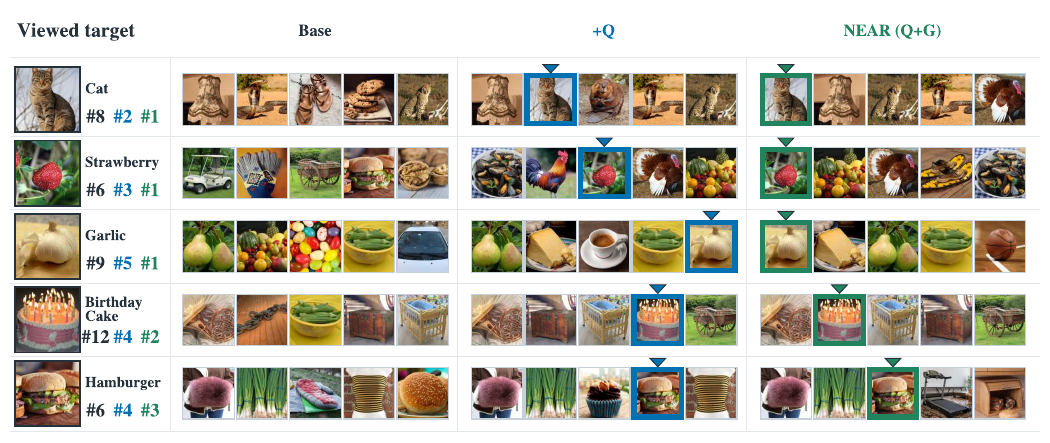}
\captionof{figure}{Representative THINGS-EEG2 retrievals at $k=4$. Colored borders mark the target among the top five under Base, $+\queryarm{}$, and \method{}.}
\label{fig:rank_flow_cases}
\end{table*}

Table~\ref{tab:main_eeg} compares the unanchored retriever, each anchoring branch, and complete \method{} under the same HCF setting. Complete \method{} exceeds the unanchored retriever at every reported repetition count on both EEG benchmarks. On THINGS-EEG2, it adds $9.3$ Top-1 points at $k=4$, and all ten participants improve at every reported repetition count. On lower-SNR Alljoined, at least 18 of 20 participants improve at every reported count, including all 20 at $k=4$. Figure~\ref{fig:rank_flow_cases} illustrates how the two branches improve individual retrievals at $k=4$: Query Anchoring raises the target rank, and complete \method{} places every shown target within the top three.

The trends match the two-term condition in Section~\ref{sec:error-analysis}. Query Anchoring contributes more at the lowest repetition counts, when query error is largest. As averaging stabilizes the query, Gallery Anchoring contributes more by improving the stable gallery margin and still helps high-$k$ concepts with limited margins. Lower SNR keeps Query Anchoring useful across Alljoined. Their combination therefore improves retrieval throughout the repetition axis.

\begin{table*}[!t]
\begin{minipage}{\textwidth}
\centering\rmfamily\footnotesize\setlength{\tabcolsep}{1.2pt}
\begin{tabular*}{\textwidth}{@{\extracolsep{\fill}}cccccccccc@{}}
\toprule
Dataset & Query estimator & $k{=}1$ & $2$ & $4$ & $8$ & $16$ & $32$ & $64$ & $80$ \\
\midrule
\multirow{4}{*}{THINGS-EEG2}
 & xDAWN   & 16.8/\best{17.9} & 28.7/\best{31.4} & 43.7/\best{48.5} & 57.6/\best{64.6} & 67.1/\best{75.7} & 72.8/\best{81.5} & 75.1/\best{84.3} & 75.9/\best{84.7} \\
 & DSS     & 17.9/\best{18.9} & 30.6/\best{33.0} & 46.6/\best{50.9} & 61.2/\best{67.3} & 70.6/\best{78.0} & 75.9/\best{83.3} & 78.8/\best{85.6} & 79.7/\best{86.2} \\
 & N2N     & 18.4/\best{19.8} & 31.3/\best{34.3} & 47.5/\best{52.5} & 62.1/\best{68.5} & 71.7/\best{79.0} & 77.2/\best{84.4} & 79.9/\best{87.1} & 80.5/\best{87.6} \\
 & Wavelet & 9.9/\best{10.5} & 17.0/\best{18.5} & 26.6/\best{29.7} & 36.6/\best{41.8} & 44.4/\best{51.8} & 49.7/\best{58.2} & 53.1/\best{62.3} & 54.0/\best{63.5} \\
\midrule
\multirow{4}{*}{Alljoined}
 & xDAWN   & 1.46/\best{1.59} & 1.99/\best{2.22} & 2.96/\best{3.36} & 4.62/\best{5.34} & 6.83/\best{7.89} & 9.46/\best{11.08} & 12.82/\best{14.34} & 13.69/\best{15.15} \\
 & DSS     & 1.41/\best{1.53} & 1.91/\best{2.15} & 2.80/\best{3.21} & 4.35/\best{5.04} & 6.63/\best{7.68} & 9.35/\best{10.37} & 11.78/\best{13.34} & 12.83/\best{14.18} \\
 & N2N     & 1.60/\best{1.85} & 2.26/\best{2.67} & 3.22/\best{3.96} & 4.84/\best{6.13} & 7.01/\best{9.01} & 9.61/\best{12.18} & 12.40/\best{15.22} & 13.43/\best{16.14} \\
 & Wavelet & 1.79/\best{1.98} & 2.50/\best{2.82} & 3.73/\best{4.29} & 5.65/\best{6.64} & 8.42/\best{9.79} & 11.68/\best{13.60} & 15.56/\best{17.72} & 16.72/\best{18.84} \\
\bottomrule
\end{tabular*}
\normalsize
\captionof{table}{Gallery Anchoring complements four query estimators. Each cell reports fixed-query / $+$\dccg{} Top-1 accuracy (\%), averaged over three seeds.}
\label{tab:query_compatibility}
\end{minipage}
\end{table*}

Besides, we test whether Gallery Anchoring depends on the query denoiser by combining it with xDAWN, DSS, Noise2Noise, and wavelet thresholding while keeping HCF fixed. Gallery Anchoring improves every repetition count on both EEG datasets for all four estimators. Thus, the gallery-side error persists across denoisers, and Gallery Anchoring corrects it without relying on one estimator.

\subsection{Gallery Anchoring Across Modalities}
\label{sec:fmri}
Table~\ref{tab:transfer} keeps each query pathway fixed and isolates Gallery Anchoring. Gallery Anchoring improves every MEG and fMRI comparison, including both NSD visual targets. All four participants in each dataset improve at every reported repetition count. Thus, the gallery-side error found in EEG also appears in other brain-imaging modalities, where the same operation improves retrieval.

\begin{table}[!htbp]
\centering\rmfamily\small\setlength{\tabcolsep}{2.5pt}
\begin{tabular*}{\columnwidth}{@{\extracolsep{\fill}}ccccc@{}}
\toprule
\multicolumn{5}{c}{THINGS-MEG (200-way)}\\
\midrule
Split & Method & $k{=}1$ & $k{=}4$ & $k{=}12$ \\
\midrule
\multirow{2}{*}{Within-subject} & Base & 7.5\smallstd{0.40} & 19.8\smallstd{0.10} & 30.9\smallstd{0.30} \\
 & $+$\dccg{} & \best{8.4}\smallstd{0.10} & \best{21.9}\smallstd{0.40} & \best{32.7}\smallstd{1.10} \\
\midrule
\multicolumn{5}{c}{NSD 7T fMRI (982-way)}\\
\midrule
Visual target & Method & $k{=}1$ & $k{=}2$ & $k{=}3$ \\
\midrule
\multirow{2}{*}{CLIP ViT-L/14} & Base & 27.5\smallstd{0.20} & 37.8\smallstd{0.10} & 43.9\smallstd{0.40} \\
 & $+$\dccg{} & \best{39.4}\smallstd{0.20} & \best{52.1}\smallstd{0.20} & \best{57.5}\smallstd{0.20} \\
\multirow{2}{*}{FARE-4 (robust)} & Base & 39.6\smallstd{0.10} & 51.6\smallstd{0.10} & 56.8\smallstd{0.20} \\
 & $+$\dccg{} & \best{54.6}\smallstd{0.20} & \best{68.6}\smallstd{0.10} & \best{74.3}\smallstd{0.20} \\
\bottomrule
\end{tabular*}
\normalsize
\caption{Gallery Anchoring improves MEG and fMRI Top-1 accuracy (\%) with each query pathway fixed.}
\label{tab:transfer}
\label{tab:fmri_main}
\end{table}

\begin{figure}[!htbp]
\centering
\includegraphics[width=\columnwidth]{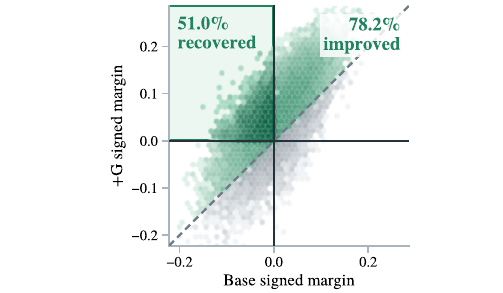}
\caption{Gallery Anchoring improves NSD gallery placement. Each point is one held-out image for one participant and seed; the upper-left region marks corrected errors.}
\label{fig:nsd_margin}
\end{figure}

The NSD gain lets us test the gallery-side mechanism directly. We average held-out repetitions to estimate each test anchor, which enters neither predictor fitting nor $+$\dccg{} retrieval. Using it as the query, we compute the signed gallery tolerance $\rho_c$ under unanchored and anchored scores. Positive values indicate correct rankings and negative values indicate errors. Figure~\ref{fig:nsd_margin} shows that Gallery Anchoring increases $\rho_c$ for $78.2\%$ of evaluations and corrects $51.0\%$ of negative unanchored values. The mean rises from $0.008$ to $0.050$, increasing in all 12 participant--seed runs. Thus, pseudo anchors improve the stable gallery margin and often correct rankings that remain wrong at the stable neural anchor.

\subsection{Stable Targets Drive Gallery Anchoring}
\label{sec:gallery-controls}
To identify what drives Gallery Anchoring, we design matched controls in Table~\ref{tab:gallery-controls}. We fix the retriever, query pathway, candidate set, and $\beta$, then replace the MLP with ridge regression, replace cosine regression with symmetric InfoNCE, or replace the stable anchor with a single-repetition neural target. Ridge uses $1.0$M rather than $8.4$M parameters, yet stays within $0.34$ points of the MLP across all four budgets. Symmetric InfoNCE retains $59$--$77\%$ of the MLP gain from $k=4$ onward, so the gain does not depend on a large nonlinear predictor or one regression loss. The target gives the clearest separation: a single-repetition target helps most at $k=1$ but falls below the unanchored retriever at $k=80$, whereas every stable-target predictor improves as the query approaches its stable reference. This reversal supports Section~\ref{sec:error-analysis}: Gallery Anchoring works by placing candidates near the stable reference approached by averaged queries.

\begin{table}[!htbp]
\centering\rmfamily\small\setlength{\tabcolsep}{1.6pt}
\begin{tabular*}{\columnwidth}{@{\extracolsep{\fill}}ccccc@{}}
\toprule
Candidate reference & $k{=}1$ & $4$ & $16$ & $80$ \\
\midrule
Base (visual)                  & 14.05 & 43.23 & 72.46 & 83.35 \\
Single-repetition target      & \best{15.78} & 45.57 & 72.41 & 81.98 \\
Stable target, InfoNCE        & 14.51 & 45.43 & 76.25 & 86.90 \\
Stable target, ridge          & 14.85 & 46.73 & 77.59 & 87.67 \\
Stable target, MLP ($+\dccg{}$) & 14.91 & \best{46.99} & \best{77.93} & \best{87.97} \\
\bottomrule
\end{tabular*}
\normalsize
\caption{Gallery controls on THINGS-EEG2. Three-seed mean Top-1 accuracy (\%) with all components except the candidate reference fixed.}
\label{tab:gallery-controls}
\end{table}

\section{Conclusion}
We studied why brain-to-image retrieval degrades with few repetitions. Target swap shows that a few-repetition query and its image can each align with a stable neural representation even when their direct alignment is weak. \method{} uses this shared anchor to denoise the query and predict a pseudo neural anchor for each gallery image.

Across EEG, MEG, and fMRI, \method{} improves retrieval with different query denoisers. Its gains follow the stable neural target across predictor classes and objectives, while signed margins confirm better gallery ordering. Thus, repeated averaging reveals a neural coordinate that can organize both sides of retrieval.

However, \method{} requires paired brain--image training data, and its participant-specific pseudo anchors must be adapted across participants and sessions. Few-repetition accuracy also remains below the high-repetition endpoint. Participant-aware prediction and stronger query estimation may further reduce the repetitions needed for reliable retrieval.

\bibliography{refs}

@inproceedings{clip,
  title={Learning Transferable Visual Models From Natural Language Supervision},
  author={Radford, Alec and Kim, Jong Wook and Hallacy, Chris and Ramesh, Aditya and Goh, Gabriel and Agarwal, Sandhini and Sastry, Girish and Askell, Amanda and Mishkin, Pamela and Clark, Jack and Krueger, Gretchen and Sutskever, Ilya},
  booktitle={International Conference on Machine Learning (ICML)},
  year={2021}
}

@inproceedings{nice,
  title={Decoding Natural Images from {EEG} for Object Recognition},
  author={Song, Yonghao and Liu, Bingchuan and Li, Xiang and Shi, Nanlin and Wang, Yijun and Gao, Xiaorong},
  booktitle={International Conference on Learning Representations (ICLR)},
  year={2024}
}

@article{muse,
  title={Mind's Eye: Image Recognition by {EEG} via Multimodal Similarity-Keeping Contrastive Learning},
  author={Chen, Chi-Sheng and Wei, Chun-Shu},
  journal={arXiv preprint arXiv:2406.16910},
  year={2024}
}

@inproceedings{atms,
  title={Visual Decoding and Reconstruction via {EEG} Embeddings with Guided Diffusion},
  author={Li, Dongyang and Wei, Chen and Li, Shiying and Zou, Jiachen and Liu, Quanying},
  booktitle={Advances in Neural Information Processing Systems (NeurIPS)},
  year={2024}
}

@article{neuroclip,
  title={{NeuroCLIP}: Brain-Inspired Prompt Tuning for {EEG}-to-Image Multimodal Contrastive Learning},
  author={Wang, Jiyuan and Zhang, Li and Lin, Haipeng and Liu, Qile and Huang, Gan and Li, Ziyu and Liang, Zhen and Wu, Xia},
  journal={arXiv preprint arXiv:2511.09250},
  year={2025}
}

@article{thingseeg2,
  title={A large and rich {EEG} dataset for modeling human visual object recognition},
  author={Gifford, Alessandro T. and Dwivedi, Kshitij and Roig, Gemma and Cichy, Radoslaw M.},
  journal={NeuroImage},
  volume={264},
  pages={119754},
  year={2022}
}

@article{thingsmeg,
  title={{THINGS}-data, a multimodal collection of large-scale datasets for investigating object representations in human brain and behavior},
  author={Hebart, Martin N. and Contier, Oliver and Teichmann, Lina and Rockter, Adam H. and Zheng, Charles Y. and Kidder, Alexis and Corriveau, Anna and Vaziri-Pashkam, Maryam and Baker, Chris I.},
  journal={eLife},
  volume={12},
  pages={e82580},
  year={2023}
}

@inproceedings{noise2noise,
  title={{Noise2Noise}: Learning Image Restoration without Clean Data},
  author={Lehtinen, Jaakko and Munkberg, Jacob and Hasselgren, Jon and Laine, Samuli and Karras, Tero and Aittala, Miika and Aila, Timo},
  booktitle={International Conference on Machine Learning (ICML)},
  year={2018}
}

@article{hcf,
  title={Aligning What {EEG} Can See: Structural Representations for Brain-Vision Matching},
  author={Tang, Jingyi and Jiang, Shuai and Su, Fei and Zhao, Zhicheng},
  journal={arXiv preprint arXiv:2603.07077},
  year={2026}
}

@article{scalinglaws,
  title={Scaling laws for decoding images from brain activity},
  author={Banville, Hubert and Benchetrit, Yohann and d'Ascoli, St{\'e}phane and Rapin, J{\'e}r{\'e}my and King, Jean-R{\'e}mi},
  journal={arXiv preprint arXiv:2501.15322},
  year={2025}
}

@article{eeg2erp,
  title={Estimating the Event-Related Potential from Few {EEG} Trials},
  author={N{\o}rskov, Anders Vestergaard and J{\o}rgensen, Kasper and Zahid, Alexander Neergaard and M{\o}rup, Morten},
  journal={Transactions on Machine Learning Research},
  year={2025}
}

@article{xdawn,
  title={{xDAWN} algorithm to enhance evoked potentials: application to brain--computer interface},
  author={Rivet, Bertrand and Souloumiac, Antoine and Attina, Virginie and Gibert, Guillaume},
  journal={IEEE Transactions on Biomedical Engineering},
  volume={56},
  number={8},
  pages={2035--2043},
  year={2009}
}

@inproceedings{makeigica,
  title={Independent Component Analysis of Electroencephalographic Data},
  author={Makeig, Scott and Bell, Anthony J. and Jung, Tzyy-Ping and Sejnowski, Terrence J.},
  booktitle={Advances in Neural Information Processing Systems (NeurIPS)},
  year={1996}
}

@article{rerp,
  title={Regression-based estimation of {ERP} waveforms: I. The {rERP} framework},
  author={Smith, Nathaniel J. and Kutas, Marta},
  journal={Psychophysiology},
  volume={52},
  number={2},
  pages={157--168},
  year={2015}
}

@article{single_trial_bci,
  title={Single-trial analysis and classification of ERP components---a tutorial},
  author={Blankertz, Benjamin and Lemm, Steven and Treder, Matthias and Haufe, Stefan and M{\"u}ller, Klaus-Robert},
  journal={NeuroImage},
  volume={56},
  number={2},
  pages={814--825},
  year={2011}
}

@article{ubp,
  title={Bridging the Vision-Brain Gap with an Uncertainty-Aware Blur Prior},
  author={Wu, Haitao and Li, Qing and Zhang, Changqing and He, Zhen and Ying, Xiaomin},
  journal={arXiv preprint arXiv:2503.04207},
  year={2025}
}

@article{bok,
  title={Rethinking Brain Decoding with {CLIP}: The Role of Adversarial Robustness},
  author={Bok, Byeongseo and Waseda, Futa and Liu, Jun and Echizen, Isao},
  journal={arXiv preprint arXiv:2607.03165},
  year={2026}
}

@inproceedings{efird,
  title={Finding Shared Decodable Concepts and their Negations in the Brain},
  author={Efird, Cory Daniel and Murphy, Alex and Zylberberg, Joel and Fyshe, Alona},
  booktitle={International Conference on Learning Representations (ICLR)},
  year={2025}
}

@inproceedings{mindeye,
  title={Reconstructing the Mind's Eye: {fMRI}-to-Image with Contrastive Learning and a Diffusion Prior},
  author={Scotti, Paul S. and Banerjee, Atmadeep and Goode, Jimmie and Shabalin, Stepan and Nguyen, Alex and Cohen, Ethan and Dempster, Aidan J. and Verlinde, Nathalie and Yundler, Elad and Weisberg, David and Norman, Kenneth A. and Abraham, Tanishq Mathew},
  booktitle={Advances in Neural Information Processing Systems (NeurIPS)},
  year={2023}
}

@inproceedings{mindeye2,
  title={{MindEye2}: Shared-Subject Models Enable {fMRI}-To-Image With 1 Hour of Data},
  author={Scotti, Paul S. and Tripathy, Mihir and Villanueva, Cesar Kadir Torrico and Kneeland, Reese and Chen, Tong and Narang, Ashutosh and Santhirasegaran, Charan and Xu, Jonathan and Naselaris, Thomas and Norman, Kenneth A. and Abraham, Tanishq Mathew},
  booktitle={International Conference on Machine Learning (ICML)},
  year={2024}
}

@article{kay2008,
  title={Identifying natural images from human brain activity},
  author={Kay, Kendrick N. and Naselaris, Thomas and Prenger, Ryan J. and Gallant, Jack L.},
  journal={Nature},
  volume={452},
  number={7185},
  pages={352--355},
  year={2008}
}

@incollection{giesbrecht2024,
  title={Electroencephalography},
  author={Giesbrecht, Barry and Garrett, Jordan},
  booktitle={Reference Module in Neuroscience and Biobehavioral Psychology},
  publisher={Elsevier},
  year={2024},
  doi={10.1016/B978-0-12-820480-1.00007-3}
}

@article{alljoined16m,
  title={{Alljoined-1.6M}: A Million-Trial {EEG}-Image Dataset for Evaluating Affordable Brain-Computer Interfaces},
  author={Xu, Jonathan and Nunes, Ugo Bruzadin and Jiang, Wangshu and Ryther, Samuel and Pringle, Jordan and Scotti, Paul S. and Delorme, Arnaud and Kneeland, Reese},
  journal={arXiv preprint arXiv:2508.18571},
  year={2025}
}

@article{nsd,
  title={A Massive {7T fMRI} Dataset to Bridge Cognitive Neuroscience and Artificial Intelligence},
  author={Allen, Emily J. and St-Yves, Ghislain and Wu, Yihan and Breedlove, Jesse L. and Prince, Jesse S. and Dowdle, Logan T. and Nau, Matthias and Caron, Brad and Pestilli, Franco and Charest, Ian and Hutchinson, J. Benjamin and Naselaris, Thomas and Kay, Kendrick N.},
  journal={Nature Neuroscience},
  volume={25},
  pages={116--126},
  year={2022},
  doi={10.1038/s41593-021-00962-x}
}

@article{chaudhary2016,
  title={Brain--Computer Interfaces for Communication and Rehabilitation},
  author={Chaudhary, Ujwal and Birbaumer, Niels and Ramos-Murguialday, Ander},
  journal={Nature Reviews Neurology},
  volume={12},
  pages={513--525},
  year={2016},
  doi={10.1038/nrneurol.2016.113}
}

@article{horikawa2013,
  title={Neural Decoding of Visual Imagery During Sleep},
  author={Horikawa, Tomoyasu and Tamaki, Masako and Miyawaki, Yoichi and Kamitani, Yukiyasu},
  journal={Science},
  volume={340},
  number={6132},
  pages={639--642},
  year={2013},
  doi={10.1126/science.1234330}
}

@article{concept2brain,
  title={Concept2Brain: an AI model for predicting neurophysiological responses to text and pictures},
  author={Santos-Mayo, Alejandro and Gilbert, Faith and Mirifar, Arash and Tebbe, Anna-Lena and Fang, Ruogu and Ding, Mingzhou and Keil, Andreas},
  journal={Nature Communications},
  year={2026},
  doi={10.1038/s41467-026-75653-x}
}

@article{wolpaw2002,
  title={Brain--computer interfaces for communication and control},
  author={Wolpaw, Jonathan R. and Birbaumer, Niels and McFarland, Dennis J. and Pfurtscheller, Gert and Vaughan, Theresa M.},
  journal={Clinical Neurophysiology},
  volume={113},
  number={6},
  pages={767--791},
  year={2002},
  doi={10.1016/S1388-2457(02)00057-3}
}

@article{mcfarland2003,
  title={Brain--computer interface ({BCI}) operation: optimizing information transfer rates},
  author={McFarland, Dennis J. and Sarnacki, William A. and Wolpaw, Jonathan R.},
  journal={Biological Psychology},
  volume={63},
  number={3},
  pages={237--251},
  year={2003},
  doi={10.1016/S0301-0511(03)00073-5}
}

@book{boyd2004convex,
  title     = {Convex Optimization},
  author    = {Boyd, Stephen and Vandenberghe, Lieven},
  publisher = {Cambridge University Press},
  year      = {2004}
}

\end{document}